\documentclass[runningheads]{llncs}
\usepackage[T1]{fontenc}
\usepackage{multirow}
\usepackage{xcolor}
\usepackage{float}
\usepackage{subcaption}
\usepackage{amsmath}
\usepackage{amsfonts}
\usepackage[algo2e,linesnumbered,ruled,vlined]{algorithm2e}
\usepackage{graphicx}
\usepackage{comment}
\usepackage{hyperref}
\usepackage{booktabs}
\usepackage[misc]{ifsym}
\begin{document}
%
\title{CRITS: Convolutional Rectifier for Interpretable Time Series Classification}

\titlerunning{CRITS}

\author{Alejandro Kuratomi\inst{1} \href{mailto:alejandro.kuratomi@dsv.su.se}{\Letter} \and Zed Lee\inst{1} \and Guilherme Dinis Chaliane Junior\inst{1} \and Tony Lindgren\inst{1} \and Diego Garc\'ia P\'erez\inst{2} \href{mailto:garciaperdiego@uniovi.es}{\Letter} } 

\authorrunning{Kuratomi et al.}

\institute{Department of Computer and Systems Sciences, Stockholm University, Borgarfjordsgatan 12, 16455 Kista, Sweden, \\ \email{{$\{$alejandro.kuratomi,zed.lee,guilherme,tony$\}$@dsv.su.se}} \and
Department of Computer Science and Engineering, University of Oviedo, Oviedo, Espa\~na, \\ \email{{garciaperdiego@uniovi.es}}}

\maketitle              

\begin{abstract}

Several interpretability methods for convolutional network-based classifiers exist. Most of these methods focus on extracting saliency maps for a given sample, providing a local explanation that highlights the main regions for the classification. However, some of these methods lack detailed explanations in the input space due to upscaling issues or may require random perturbations to extract the explanations. We propose Convolutional Rectifier for Interpretable Time Series Classification, or CRITS, as an interpretable model for time series classification that is designed to intrinsically extract local explanations. The proposed method uses a layer of convolutional kernels, a max-pooling layer and a fully-connected rectifier network (a network with only rectified linear unit activations). The rectified linear unit activation allows the extraction of the feature weights for the given sample, eliminating the need to calculate gradients, use random perturbations and the upscale of the saliency maps to the initial input space. We evaluate CRITS on a set of datasets, and study its classification performance and its explanation alignment, sensitivity and understandability.\footnote{This paper was presented at the 6th eXplainable Knowledge Discovery in Data Mining, XKDD, Workshop of the European Conference on Machine Learning and Principles and Practice of Knowledge Discovery in Databases, ECML-PKDD 2024, in Vilnius, Lithuania.}


\end{abstract}

\section{Introduction}
\label{sec:introduction}


Time Series (TS) data is any data containing a set of time-evolving variables. A Multivariate TS (MTS) is a set of TS containing more than one time-evolving variable, whilst a Univariate TS (UTS) is a TS containing only one time-evolving variable. TS are increasingly used in different high-stake decision-making scenarios where machine learning models are used to perform classification tasks, i.e., TS Classification (TSC) \cite{theissler2022explainable,ruiz2021great,ismail2019deep,rojat2021explainable}. Among these scenarios, we may find automotive systems, such as vehicle localization \cite{kuratomi2020prediction}; medical applications, such as heartbeat anomaly detection \cite{maletzke2013time}; and predictive maintenance, such as engine operation anomaly detection \cite{kiangala2020effective}.

The increasing complexity of the machine learning models has earned them the \emph{``black-box''} label, i.e., models with low transparency and low interpretability. Here, we understand \textit{interpretability} as a passive characteristic of models that allows users to understand how the given inputs are processed towards the resulting outputs \cite{theissler2022explainable,rudin_2,rojat2021explainable,biran2017explanation,molnar}. Specifically for TSC, understanding the inner mechanisms of the model can be achieved by estimating the relevance of each time step in each variable for the model's classification output. A widely used \textit{explanation} is an estimation of such relevance.

Measuring the quality of the explanations is challenging and there is no consensus on the best method for evaluation \cite{kuratomi2022juice,jia2019improving}. Furthermore, it is harder to attain explanations in TS when compared to other kinds of data, such as tabular, image, and text \cite{rojat2021explainable}. This is due to the fact that we do not usually get a visual representation of time-dependent phenomena. For example, images of cats or dogs instantly reveal the features that are relevant to classify them (ears, nose, etc.), but sounds of notes played on a piano and a violin hardly allow us to tell which features make the sound unique to each. By analogy, recordings of speech or other time-dependent data are hard to classify and interpret without additional insights \cite{rojat2021explainable}. For UTS, the signal plot may help to understand its behavior, but the problem becomes harder with MTS.

Convolutional network-based models have been widely employed for TSC and specially for MTS Classification. These models convolve the input feature values to learn patterns that enable models to distinguish among the different classes. However, these convolutions considerably decrease the models interpretability, since the identified patterns cannot be easily backtracked to the original input space \cite{zhou2016learning,selvaraju2017grad}. Correspondingly, a set of algorithms have been developed to improve the interpretability of these convolutional-based models and compute the input relevance.

The first interpretability algorithm for these models is called Class Activation Map (CAM) \cite{zhou2016learning}. Originally designed to explain convolutional neural networks in image classification, CAM obtains a \textit{saliency map} (a relevance matrix) that highlights the feature relevances through time for an input instance. However, CAM has some disadvantages: (1) it constrains the models architecture by discarding the fully-connected layers after the pooling layer, which eliminates deep learning parts in the model (increasing interpretability) but decreases its classification performance; (2) the saliency maps do not provide fine-grained explanations due to the upsampling required to translate them from the convolved space to the original input space. The latter is a problem specially when detailed explanations are required, since the initial downsampling caused by the convolutional and pooling layers erases part of the input details. Then, when trying to upsample them to regain this information, some artifacts are introduced that distort the feature relevance and cause a loss of explanation sharpness \cite{chollet2021deep,selvaraju2017grad}.

\sloppy
The first issue was solved by Selvaraju et al. with the Gradient-weighted Class Activation Mapping (Grad-CAM) model \cite{selvaraju2017grad}. Grad-CAM allows fully-connected layers after the pooling layer, increasing the classification performance and improving the models interpretability. Grad-CAM uses backpropagated gradients to obtain more detailed explanations up to the convolutional layers, however, it still uses bilinear interpolation to upsample these explanations back into the original input space, which blurs the explanation \cite{chollet2021deep,widdicombe2021gradient}.

CAM-based methods are considered \textit{post-hoc} interpretability, since the explanations are computed from trained classifiers. Other post-hoc methods are based on input gradients with respect to the model's ouput. Smilkov et al. studied the problem of noisy gradient-based saliency map explanations and proposed SmoothGRAD \cite{smilkov2017smoothgrad}, a post-hoc interpretability method that uses random perturbations to sharpen the obtained explanations and make them more robust to noise and variations in the partial derivates when calculating gradients \cite{smilkov2017smoothgrad}. However, SmoothGRAD has two user-defined parameters: the noise level and the number of perturbations to use. This creates potentially different explanations for the same instance even when running it with the same parameters. Other post-hoc methods, such as the Linear Interpretable Model-agnostic Explanations (LIME) approach \cite{ribeiro2016should}, use the weights of linear models trained around an input instance. Although LIME is a model-agnostic method, it suffers from the same instability problem as gradient-based methods.

Contrary to post-hoc methods, intrinsically interpretable methods inherently reveal the input-output relations involved in the model's output decision without actively computing an explanation. Examples of such methods include linear regression, decision trees (with decision paths from root to leaf), and neural networks that use only Rectified Linear Unit (ReLU) activations, also known as Rectifier Networks (RN) \cite{molnar,sudjianto2020unwrapping}. In these approaches, their inner mechanisms can be tracked to extract input relevance.

We propose a model called Convolutional Rectifier for Interpretable Time Series classification (CRITS) for TSC that provides sharp, detailed saliency map explanations \textit{without} calculating gradients, upsampling from the convolved space, adding noise or using random perturbations. The algorithm is intrinsically interpretable, it is based on the unwrapper method proposed by Sudjianto et al. \cite{sudjianto2020unwrapping} and is adapted to explain convolutional-based models. CRITS has three components: (1) a convolutional layer followed by (2) a max-pooling layer that connects to (3) a RN. CRITS explanations are extracted by first unwrapping the RN to find the weights at the max-pooling activations, then, we trace the max-pooling activations back to the relevant feature maps and extract the weights of the times and features in the TS, as used by the classifier. This architecture obtains the set of TS input feature relevance values through the exact weights used by the model for each feature value at each time step to make the decision. We demonstrate CRITS classification performance on different public datasets, both UTS and MTS, and evaluate its explanations through three measures: the change in the classifiers score when perturbing the relevant features indicated by the explanation, the sensitivity of the feature relevances to changes in the input feature values and the level of explanation sparsity as a proxy for its simplicity and understandability.

Section \ref{sec:related_work} of this paper briefly recounts the relevant related work, providing details about TS explainability algorithms. Then, section \ref{sec:method} presents the methodology applied, describing the mathematical notation, the CRITS algorithm and the experimental setup. Section \ref{sec:empirical} provides the results and benchmark of the classification performance and the explanation assessment of CRITS. Finally, section \ref{sec:conclusions} presents the conclusions and future work.

\section{Related work}
\label{sec:related_work}

The RandOm Convolutional KErnel Transform (ROCKET) algorithm \cite{dempster2020rocket} uses a single convolutional layer with several randomly selected filters, followed by a pooling method that extracts the maximum value and the fraction of positive values per filter, and a linear classifier model. Albeit its low interpretability, it has an outstanding generalizability, providing remarkable accuracy throughout different datasets \cite{dempster2020rocket}. Its high classification performance (and its low interpretability) comes mainly from the large amount of random convolutional kernels and the usage of the proportion of positive values as a feature.

According to Theissler et al. \cite{theissler2022explainable}, the explanations for TS datasets can be categorized into three groups: time-point-based, which highlight the most important time steps for the models decision; subsequences-based, which underline the most relevant subsequences of each variable; and instance-based, which indicate how important each feature is for a given instance \cite{theissler2022explainable,ruiz2021great,ismail2019deep,rojat2021explainable}.

Among time-point-based explanation methods, we find the \textit{attention} mechanism, which can learn the input features importance as the model is trained. However, this method demands the training of additional model components, increasing its complexity \cite{theissler2022explainable} and it may provide unstable explanations, i.e., weights that can be altered (for example, with adversarial attacks), weights that are dissimilar and leading to the same output, or weights that are actually not the ones used by the model \cite{theissler2022explainable,burstein2019proceedings,hsu2019multivariate,hsu2019multivariate}.

Additionally, saliency maps are visualizations of \textit{attribution} explanation methods, which also make part of time-point-based explanation methods. The CAM \cite{zhou2016learning} and Grad-CAM \cite{selvaraju2017grad} models are part of the attribution methods. CAM uses a combination of CNN layers followed by a Global Average Pooling (GAP) layer connected to a fully-connected layer. In this architecture, the activation is back-propagated with the weights from the fully-connected layer to the last convolutional layer, and a saliency map is obtained which approximately weights the input features. The main difference with Grad-CAM is that Grad-CAM may use any network after the pooling layer and backpropagates the gradients to obtain a better classification performance and explanation. However, these saliency maps lack the sharpness required to describe the exact input feature weights used by the model \cite{zhou2021salience,theissler2022explainable,chollet2021deep,widdicombe2021gradient}, and could prove hard to understand when assessing each specific point in time \cite{theissler2022explainable}. Additionally, the usage of average pooling, albeit identifying all the relevant regions of an input, decreases the explanations simplicity. A max-pooling layer highlights the most discriminative input features \cite{zhou2016learning}.

Among other attribution methods, we find DeepSHAP and GradientSHAP, which are based on Shapley values \cite{veerappa2022validation}. Shapley values are the contributions of each feature value towards the models outcome. In order to obtain these, different coalitions of feature values must be formed and their output impact estimated \cite{molnar}. In the case of DeepSHAP, an attribution algorithm known as DeepLIFT \cite{shrikumar2017learning}, which uses difference backpropagation, is used to approximate the Shapley values using different reference instances. For GradientSHAP, the Shapley values are approximated by averaging integrated gradients across random paths from different reference instances \cite{lundberg2017unified}.

Among the subsequence-based explanations are \textit{shapelets} which are sets of TS segments either sampled from the dataset or synthetically formed \cite{gupta2020approaches,theissler2022explainable,dorle2020learning}. A model is built to classify the instances based on their similarity towards the class discriminative shapelets. Inside the subsequence and instance-based interpretability methods, \textit{prototypes} are similar to shapelets, and are usually class representative instances used to classify a TS using a distance measure \cite{theissler2022explainable,zhang2020tapnet}.

Shapelets and prototypes, counterintuitively, do not necessarily present high interpretability \cite{gupta2020approaches,zhang2020tapnet}. Given a set of shapelets or prototypes, the relevant shapes and characteristics observed in them are often not easily understandable patterns and experts may be needed to comprehend them \cite{theissler2022explainable}. The problem is noteworthy when the shapelet or prototype is synthetically generated, since this generation usually requires a complex model, leading to an undesired loop of interpretability (an opaque model to explain an opaque model requires an explainer \cite{rudin_2}) and the generated sequences could be even less interpretable \cite{theissler2022explainable}.

Regarding neural networks interpretability, Sudjianto et al. \cite{sudjianto2020unwrapping} proposed an unwrapper feed-forward method to extract the weights used by a RN for a given instance. The advantage of these is that they are the exact weights the model uses to make the decision, i.e., they are not approximations. The algorithm is designed for fully-connected RNs and it does not work with other architectures, such as a convolutional neural network (CNN), \cite{sudjianto2020unwrapping}, since it requires a continuous set of linear transformations, which the convolutional kernels interrupt.

Finally, assessing the explanation quality is a problem \cite{kuratomi2022juice,jia2019improving}. One way is to analyze the change in the models prediction scores with respect to changes in an instances input feature values. The idea is to perturb the input feature values with respect to their relevance, as indicated by the explanation. The reasoning is: if the explanation is accurate, then the sensitivity of the models performance with respect to these relevance-guided changes should be high, i.e., the models prediction scores should change significantly when compared to changing irrelevant features \cite{schlegel2019towards}. There are several ways to perturb a TS instance. \cite{schlegel2019towards} propose four TS perturbation methods to evaluate the accuracy of the explanations. The perturbations proposed are: (1) \textit{zero}, which replaces feature values with zero; (2) \textit{inverse}, which replaces feature values with the maximum value of the TS minus the current feature value; (3) \textit{swap}, which flips a subsequence of feature values in time (chronologically reversed); and (4) \textit{mean}, which replaces a subsequence of values for their mean. Another way is to evaluate the sensitivity with respect to small changes in the input \cite{smilkov2017smoothgrad}: small changes in the input should not significantly change the extracted explanation. Consequently, large input changes should significantly change the explanation. Finally, the explanation simplicity or sparsity is important: simpler and shorter explanations are preferred \cite{miller2019explanation}.

Summarizing, interpretability in TS faces some challenges: firstly, the weights obtained through time-point-based explanations, such as attention and saliency maps, may not represent the exact input feature weights used by the model. Additionally, the random perturbation-based post-hoc explainers, like LIME or SmoothGRAD, add additional user-defined parameters that may highly influence the explanation output. With subsequence or instance-based explanations, there is a risk of finding patterns that are not interpretable. Therefore, we bring in intrinsic interpretability by generating an architecture that is able to trace back the \textit{exact} weights as used by the model from the last output neuron, passing through a set of fully-connected layers, a max-pooling layer, the feature-extracting convolutional layers, to the input TS and attain local interpretability. 

\section{Methodology}
\label{sec:method}

In this section, we first present the required concepts to understand the CRITS algorithm. Then, we present the CRITS approach and discuss the process by which it extracts the input feature weights as an explanation. Finally, we describe the experimental setup before proceeding to the empirical evaluation. The implementation can be found in the GitHub repository \footnote{\url{https://github.com/alku7660/CRITS}}.

\subsection{Preliminary concepts}

We define a binary classification task on a TS dataset (which could be univariate or multivariate) $\mathbf{D} = (\mathbf{X}, \mathbf{Y})$ where $\mathbf{X} = \{\mathbf{x}^1, \mathbf{x}^2, \dots, \mathbf{x}^N\}$ and each instance $\mathbf{x}^i$ may be defined as a set of observations $\mathbf{x}^i = \{x^i_{11}, x^i_{21}, \dots, x^i_{mT-1}, x^i_{mT}\}$ of $m$ variables over $T$ time steps, $\mathbf{X} \in \mathbb{R}^{N \times T \times m}$, and $\mathbf{Y} = \{y^1, y^2, \dots, y^N\}, \mathbf{Y} \in \{0, 1\}^N$ a set of class labels. We define a classifier function $f$ with a final sigmoid activation function, trained so that $f(\mathbf{X}) = \mathbf{\hat{Y}}$ and the error between $\mathbf{\hat{Y}}$ and $\mathbf{Y}$ is minimized. 

Additionally, we refer to $\mathbf{w}^i$ as the set of feature weights for every $\mathbf{x}^i \in \mathbf{X}$ that relates input elements to output a class label, $\mathbf{w}^i = \{w^i_{11}, w^i_{21}, \dots, w^i_{mT-1}, w^i_{mT}\}$. In case of linear classifiers, $\mathbf{w}^i$ are used by the classifier $f$ to output a class label ($y^{i}=\mathbf{w}^{iT}\mathbf{x}^i$), while in non-linear classifiers, such as CNN models, $\mathbf{w}^i$ cannot be directly extracted. In these cases, $\mathbf{w}^i$ are usually computed by local post-hoc models, such as CAM-based methods. We also denote the relevance value as $\mathbf{r}^i = \mathbf{w}^i \odot \mathbf{x}^i = \{w^i_{11}x^i_{11}, w^i_{21}x^i_{21}, \dots, w^i_{mT-1}x^i_{mT-1}, w^i_{mT}x^i_{mT}\}$, where $\odot$ is the element-wise multiplication operator. $\mathbf{r}^i$ indicates the value of each of the terms used in the sigmoid activation function in the output of the classifier $f$.

Finally, we define the function $p_f(\mathbf{x}^i)$ as the probability of the classifier $f$ of returning 1 as the label for instance $\mathbf{x}^i$, i.e., the prediction probability function and \textit{explainer} function $\mathcal{E}(f,\mathbf{x}^i)$ that outputs the explanation of model $f$ for the prediction of instance $\mathbf{x}^i$.

\subsection{Convolutional Rectifier for Interpretable Time Series Classification (CRITS)}

\begin{figure}
    \centering
    \includegraphics[trim={1cm 1.25cm 0.75cm 1.2cm},clip,width=0.9\linewidth]{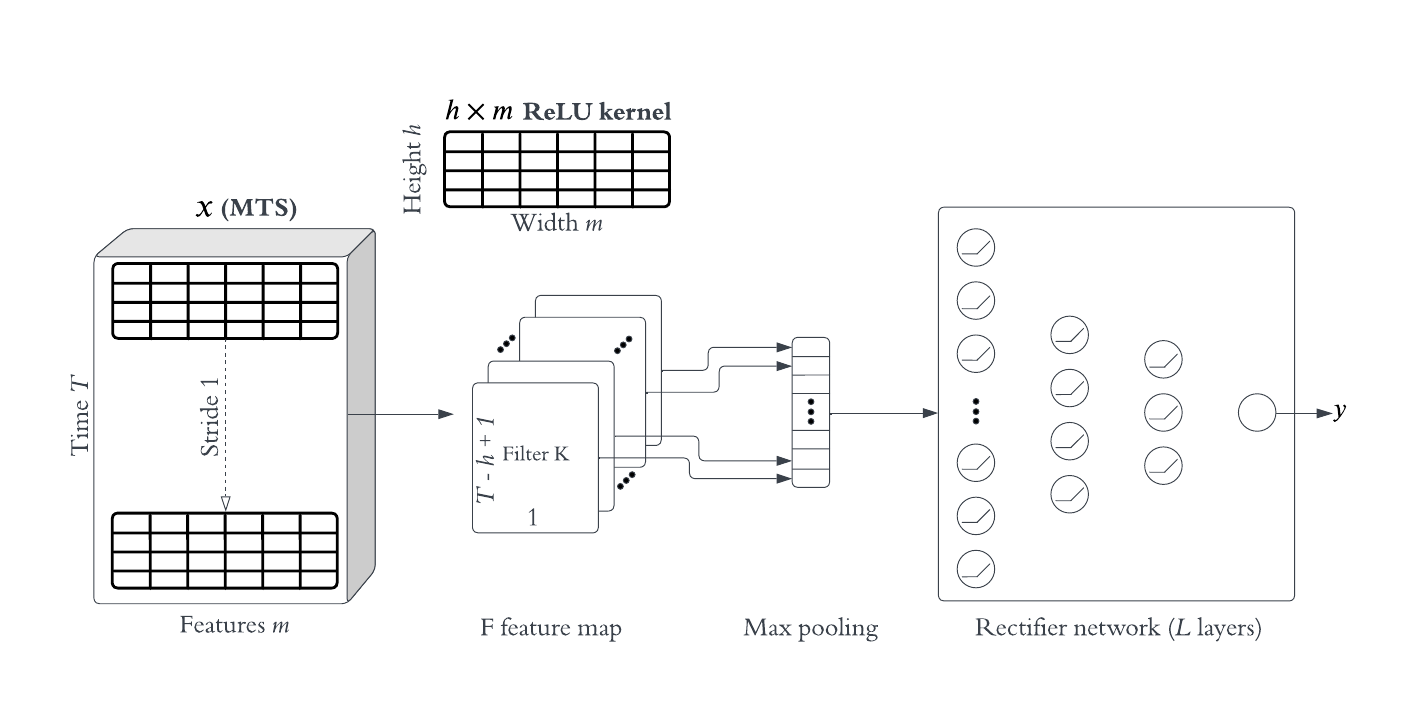}
    \caption{CRITS model: a convolutional layer outputting a feature map $F$ composed of $K$ filters, followed by a max-pooling layer and a RN with $L$ layers.}
    \label{fig:crits}
\end{figure}

Fig. \ref{fig:crits} shows the CRITS architecture. The normalized MTS $\mathbf{x}^{i}$ is used as input. The model uses first a \emph{convolutional layer} which applies $K$ kernels of size $h \times m$ throughout all the input with a stride of $1$. The output feature map $\mathbf{F}^{i} \in \mathbb{R}^{(T-h+1)\times K}$ of the convolutional layer is fed into a \emph{global max-pooling layer}. The output of the global max-pooling layer is then used as the input to a RN, where the last, output neuron has a sigmoid activation function.

\subsection{Extracting input feature weights from CRITS}

In order to obtain the weights $\mathbf{w}^i$ and relevance values $\mathbf{r}^i$, we input an instance $\mathbf{x}^i$ into the model and follow two steps: (1) Unwrap the RN (the last component of the model) to obtain the set of weights $\mathbf{w}^i_{K}$, a step that was already developed as the unwrapper for RNs in \cite{sudjianto2020unwrapping}; and (2) apply a deconvolution from the filters selected by the global max-pooling layer to the initial TS $\mathbf{x}^i$ input features, and weight these with the $\mathbf{w}^i_{K}$ weights from the RN, which is the extension proposed for convolutional and pooling layers in this type of classifiers.

\subsubsection{Unwrapping the RN.}

Unwrapping the RN is done by following Eqs. \ref{eq:unwrapper_coef} and \ref{eq:unwrapper_bias}, as described by Sudjianto et al. \cite{sudjianto2020unwrapping}.

\begin{align}
    \label{eq:unwrapper_coef}
    \mathbf{w}^i_{K} &= \prod_{l \in L}\mathbf{w}_{(L+1-l)}\mathbf{p}^i_{(L+1-l)}\mathbf{w}_0\mathbf{x}^i, \\
 \label{eq:unwrapper_bias}
    b^i_{K} &= \sum_{n \in L}\prod_{l \in L}\mathbf{w}_{(L+1-l)}\mathbf{p}^i_{(L+1-l)}\mathbf{b}_{n-1} + \mathbf{b}_L,
\end{align}

where $\mathbf{w}_l$ and $\mathbf{b}_l$ correspond to the weights and biases, respectively, at the $l$ RN layer. $\mathbf{p}^i_l$ corresponds to the activation pattern of the layer $l$ given the input from the max-pooling layer. The activation pattern $\mathbf{p}^i_l$ is a vector of zeros and ones, indicating a one whenever the input is positive (ReLU outputs the input) and a zero whenever the input is negative (ReLU outputs a zero). This creates a linear model $z^{i}= \mathbf{w}_K^{i} \textrm{maxpool}(\mathbf{F}^{i}) + b_k^{i}$ that matches the decision output $z^{i}$ ---before the sigmoid activation--- with the max-pooling layer output values and allows the RN to have $L$ layers without losing intrinsic interpretability. 

The magnitude of $\mathbf{w}^i_{K}$ is the output relevance of the maximum value from each learned filter $K$. To be able to trace back from the max pooling layer all the way to the initial MTS $\mathbf{x}^i$ input features, the weights $\mathbf{w}_K^{i}$ should be back propagated by applying a \emph{deconvolution operation}.

\subsubsection{Deconvolution towards the input features.} 
\begin{figure}[!t]
    \centering
    \includegraphics[trim={1.1cm 1.25cm 0.8cm 1cm},clip,width=0.9\linewidth]{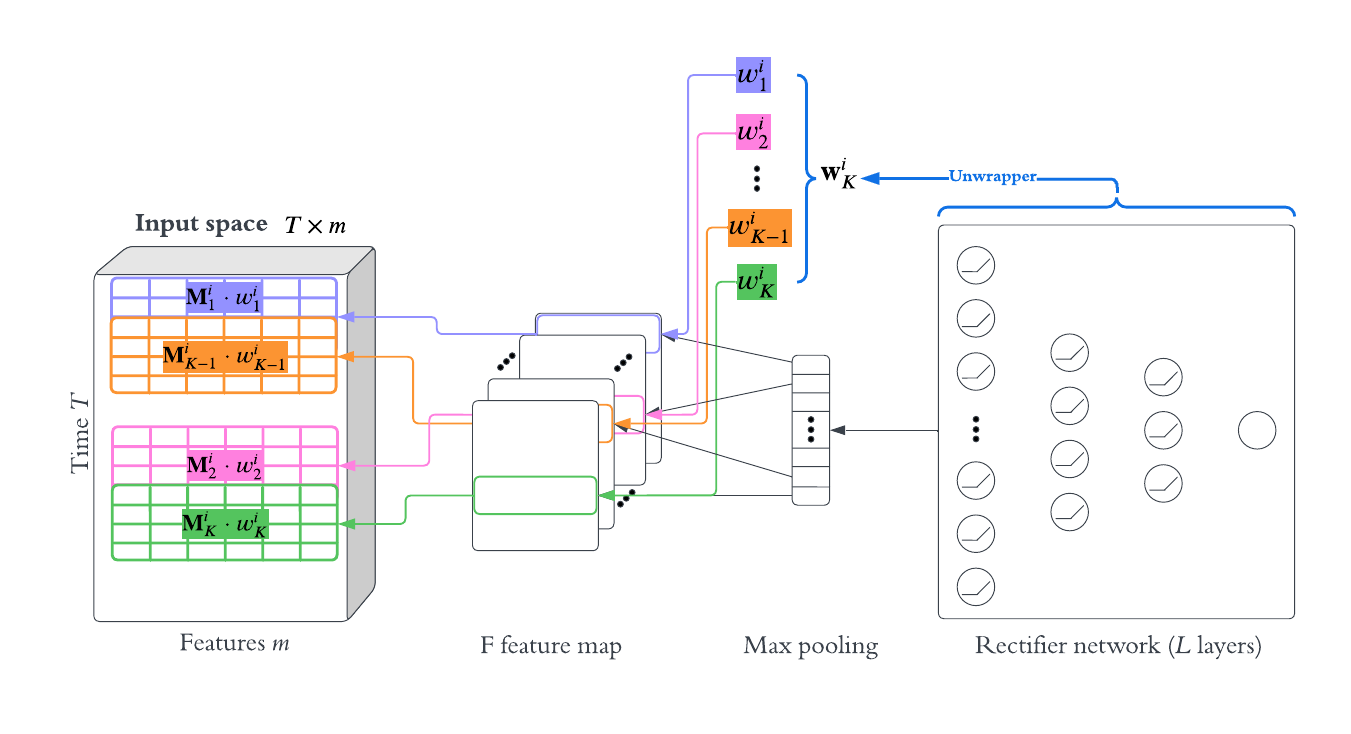}
    \caption{Deconvolution process. The unwrapper (blue arrow) obtains the $\mathbf{w}^i_{K}$ max-pooling weights. These weights are traced back to the kernels containing the maxima (each color represents a kernel weight). The weights of each of these kernels are arranged into a mask $\mathbf{M}^i_k$ of size $T \times m$ and then weighted by the corresponding $\mathbf{w}^i_{K}$ values and added together.}
    \label{fig:deconvolution}
\end{figure}

We extend the unwrapper proposed by Sudjianto et al. and follow a similar logic for the convolution and pooling layers. The convolution and max-pooling operations of CRITS are unwrapped by a \emph{deconvolution} and an \emph{unpooling} operation \cite{zeiler2014visualizing}, respectively. The global max-pooling layer implies that the network considers an input portion of size $h \times m$ --attached to the convolution step with maximum activation-- per filter, bringing sparsity. In order to invert this operation, the position of the feature map associated with the maxima is stored and the kernel's weights are then projected into the corresponding input space (see Fig. \ref{fig:deconvolution}). This process results in a mask of the same dimension as the input $M^{i}_k \in \mathbb{R}^{T \times m}$ per filter, being all its elements zeros except $(h \times m)$ elements corresponding to the projected kernel's weights. A sum of all $M^{i}_k$ weighted by the unwrapped weights from the RN $\mathbf{w}^{i}_K$ is used to get the final weights. Eq. \ref{eq:deconvolution} shows the computation of the $\mathbf{w}^i$ weights using the mask and the unwrapped RN max-pooling weights $w^i_k$.

\begin{equation}
    \label{eq:deconvolution}
    \mathbf{w}^i = \sum_{k=1}^{K}\mathbf{M}^{i}_{k} \cdot w^{i}_k
\end{equation}

The weights $\mathbf{w}^i$, biases $ b^i_{K}$ and input $\mathbf{x}^{i}$ relate to the output $z^{i}$ as follows:
\begin{equation}
    z^{i} = \mathbf{w}^{i}\mathbf{x}^{i} + b_K^{i},
\end{equation}

where the $z^{i}$ terms form a relevance map $\mathbf{r}^{i} \in \mathbb{R}^{T \times m}$ of the same dimensionality as the input, computed as $\mathbf{r}^i = \mathbf{w}^i \odot \mathbf{x}^i$. These are the terms used as input for the sigmoid function in the last output neuron.

\subsection{Experimental setup}
\label{sec:experimental}
The goal of these experiments is to evaluate the CRITS architecture at classification and extracting the explanations intrinsically. To this objective, we train CRITS using three UTS and three MTS TSC datasets\footnote{ Available at \url{https://www.timeseriesclassification.com/dataset.php}} described in Table \ref{tab:datasets}. These datasets are GunPointMaleVersusFemale (GunPoint), SharePriceIncrease, Strawberry, Blink, SCP1 and Heartbeat.

\begin{table}[]
    \centering
    \begin{tabular}{llccc}
    \toprule
    \textbf{Dataset} & \textbf{Description} & \textbf{$m$} & \textbf{$T$} & \textbf{$N$} \\
    \midrule
    GunPoint & Male and Female in gun and point movement & 1 & 150 & 451 \\
    \midrule
    SharePriceIncrease & Share price increase prediction & 1 & 60 & 1930  \\ 
    \midrule
    Strawberry & Fruit spectroscopy for strawberries and other fruits & 1 & 235 & 983 \\
    \midrule
    Blink & Long and short eye blinking EEG signals & 3 & 510 & 1500\\
    \midrule
    SCP1 & Self-regulation of slow cortical potentials & 6 & 896 & 561 \\
    \midrule
    Heartbeat & Heart pathology based on sound recording & 61 & 405 & 409 \\
    \bottomrule
    \end{tabular}
    \caption{Description of the datasets used in the empirical evaluation.}
    \label{tab:datasets}
\end{table}

For each of the datasets, a hyperparameter random search is performed to find the best parameters for CRITS. The details of this search and the best classifier parameters per dataset are given in Appendix \ref{appendix:hyperparameters}.

We evaluate the performance of CRITS from two perspectives: the classification performance and the quality of the explanations. For the classification performance, we use the F1-score for each of the datasets and compare it with the F1-score obtained by the ROCKET algorithm. In terms of explanation quality, we have three main sub-metrics: (1) the change in the models prediction score due to changes in high-relevance input elements (here referred to as the explanations \textit{alignment} with the model); (2) the change in the explanations extracted due to changes in the input feature values (here referred to as the explanations \textit{input-sensitivity}) and (3) the \textit{sparsity} of the explanations  \cite{miller2019explanation}. 

Before proceeding, we introduce some additional concepts. Let us define a perturbation of instance $\mathbf{x}^i$ as instance $\mathbf{\hat{x}}^i$, and the function $p_f(\mathbf{x}^i)$ as the probability of the classifier $f$ of returning 1 as the label for instance $\mathbf{x}^i$, i.e., the prediction probability function. Finally, let us define an \textit{explainer} function $\mathcal{E}(f,\mathbf{x}^i)$ that outputs the explanation of model $f$ for the prediction of instance $\mathbf{x}^i$.

Regarding the sub-metrics for the assessment of explanations, the higher the alignment the better: a high alignment indicates that the explanations are highlighting the features considered to be relevant by the model, since changes on these features lead to a higher change on the model's decision outputs. The input-sensitivity is better if it increases as the magnitude of the input changes increase, i.e., it is low for small input changes and high for larger input changes. Sparsity is better if it is lower, meaning that the explanations have a low number of highly relevant features, reflecting simpler and easier to understand explanations.

The proposed alignment metric is inspired in perturbation-based metrics suggested in \cite{schlegel2019towards}. More specifically,  alignment is computed as:

\begin{equation} \label{eq:alignment}
    A = RMSE(p_f(\mathbf{x}^i) - p_f(\mathbf{\hat{x}}^i)),
\end{equation}

\noindent where $\mathbf{\hat{x}}^i$ is the perturbed instance corresponding to the input instance $\mathbf{x}^i$, with perturbations on relevant time steps and features. $A$ is measured as the RMSE between the probability outputs of the classifier $f$ for the $\mathbf{x}^i$ and the $\mathbf{\hat{x}}^i$ instances. Regarding the perturbations, we use the four perturbations proposed by \cite{schlegel2019towards} to evaluate the accuracy of the explanations, namely \textit{zero}, \textit{inverse}, \textit{swap} and \textit{mean}.
For each of the perturbation methods, the metric $A$ is computed for $50$ test samples. This process is repeated $5$ times, selecting randomly different test sets. The obtained results are then statistically evaluated in Section \ref{sec:explanation_quality}. Eq. \ref{eq:input-sensitivity} shows the estimation of the input-sensitivity measure.

\begin{equation} \label{eq:input-sensitivity}
    IS = RMSE(\mathcal{E}(f,\mathbf{x}^i), \mathcal{E}(f,\mathbf{\hat{x}}^i)),
\end{equation}

\noindent where the $IS$ measure estimates the difference between the explanations extracted on the original instance and its perturbation. In the case of the sensitivity evaluation $\mathbf{\hat{x}}^i$ instances are obtained adding Gaussian noise to the original input. The noise magnitude is defined via its standard deviation.

The Sparsity is measured as the fraction of input features having a weight with a magnitude higher than 0.01, as shown in Eq. \ref{eq:sparsity}.

\begin{equation} \label{eq:sparsity}
    S = \frac{\lvert \{\mathbf{w}^i | |\mathbf{w}^i| > 0.01\} \rvert}{m \times T},
\end{equation}

\noindent where $S$ is the sparsity. The lower the $S$ is, the lower the amount of relevant features and the simpler to understand the explanation is \cite{miller2019explanation}. We obtain four different explanations for each test instance: the original intrinsic explanations from CRITS, and an explanation from the DeepSHAP, GradientSHAP and the SmoothGRAD post-hoc explainer methods. We then assess how each of these explanations perform in terms of alignment, input-sensitivity and sparsity and compare them to the CRITS intrinsic explanations.

\section{Empirical Evaluation}
\label{sec:empirical}

We present here the results of the CRITS model in terms of classification performance and explanation quality. Appendix \ref{appendix:qualitative_results} presents qualitative examples of the saliency maps for each of the explanation methods for some instances of each dataset.

\subsection{Classification performance}
\label{sec:classification}

The performance of CRITS is shown in Table \ref{tab:classification} in terms of F1-score, showing a high classification performance comparable to ROCKET. 

\begin{table}[]
    \centering
    \begin{tabular}{lcc}
        \textbf{Dataset} & \textbf{CRITS} & \textbf{ROCKET} \\ 
        \midrule
        GunPoint & 0.99 & 1.00 \\
        \midrule
        SharePriceIncrease & 0.62 & 0.66 \\
        \midrule
        Strawberry & 0.99 & 0.98 \\
        \midrule
        Blink & 1.00 & 1.00 \\
        \midrule
        SCP1 & 0.88 & 0.85 \\
        \midrule
        Heartbeat & 0.72 & 0.74 \\
        \bottomrule
    \end{tabular}
    \caption{F1-score for each dataset.}
    \label{tab:classification}
\end{table}

\subsection{Explanation quality}
\label{sec:explanation_quality}

We present here the results of CRITS with respect to alignment, input-sensitivity and sparsity. Consider also that the explanations weights (saliency maps) from CRITS are the actual weights used by the CRITS model.

\subsubsection{Alignment.} 
\begin{figure}[t!]
    \centering
    \includegraphics[trim={0.15cm 0.2cm 0.4cm 0cm},clip,width=0.92\linewidth]{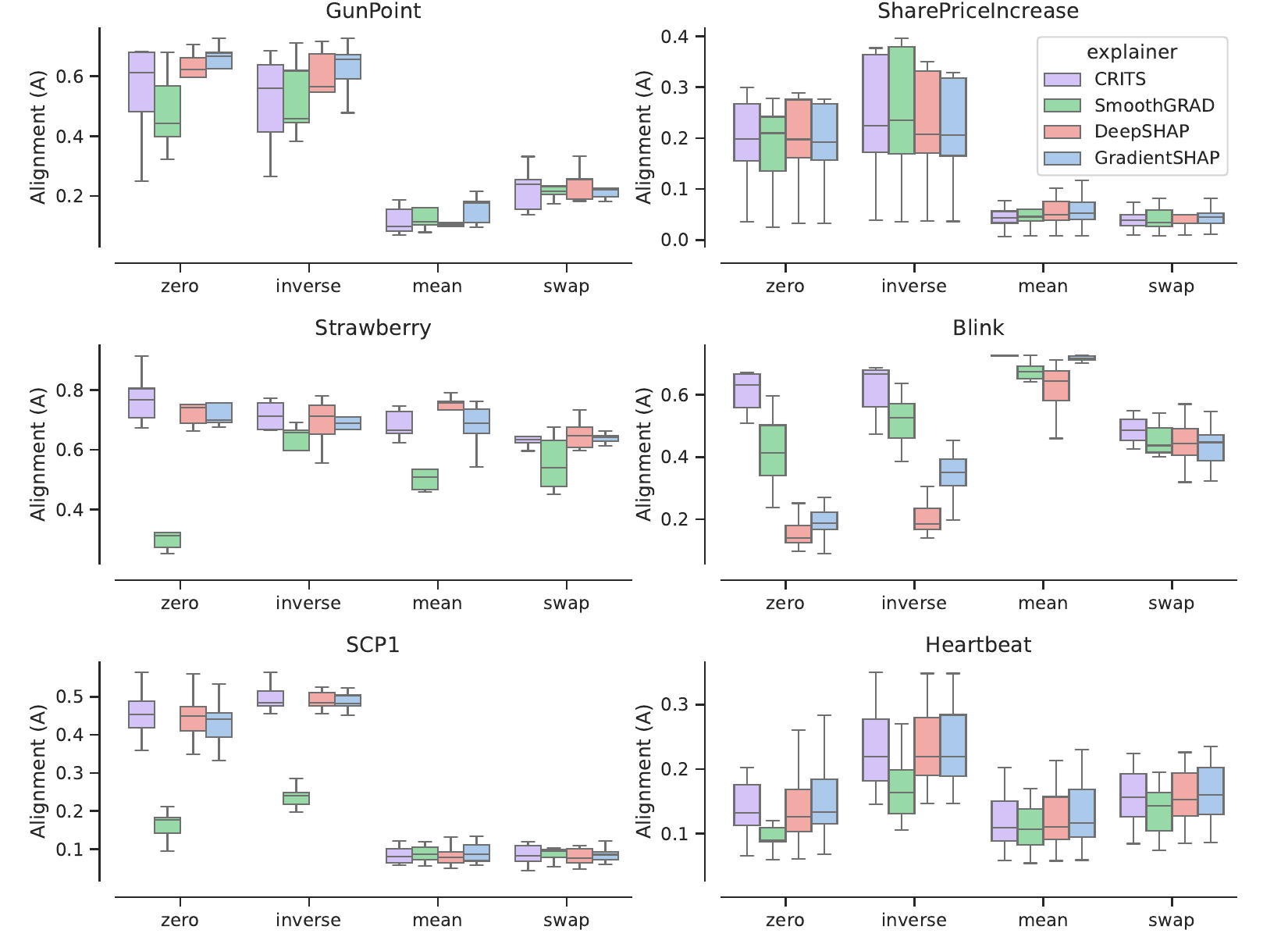}
    \caption{The explanations alignment for each datasets and perturbation form.}
    \label{fig:alignment}
\end{figure}

Figure \ref{fig:alignment} is the boxplot of the alignment for each of the explanations, datasets, and each of the perturbation methods computed across several trials (see Section \ref{sec:experimental}). Ideally, the alignment should be as high as possible.

In theory, the model should change its prediction score when the weights highlighted as relevant by CRITS are changed, since these are the exact weights used by the model. In general, the alignment of CRITS is good, albeit the behavior of the alignment is strongly dependent on the dataset and perturbation method. In the zero and inverse perturbations, CRITS presents the highest alignment in Strawberry, SCP1, Heartbeat and Blink. In the GunPoint and SharePriceIncrease, CRITS has either the second or third place in alignment in these two perturbation methods. In Blink, CRITS has the highest alignment in all of the perturbation methods. CRITS does not seem to perform well with the mean perturbation method, positioning last in the GunPoint, SharePriceIncrease and SCP1 datasets for this specific method, but this behavior seems to be truth for the mean and swap methods, given that these two usually seem to present a lower level of perturbation magnitude and the alignment for all explainers seems lower than for the zero and inverse perturbation modes.

\subsubsection{Input-sensitivity.} 
\begin{figure}
    \centering
    \includegraphics[trim={0.15cm 0.2cm 0.4cm 0cm},clip,width=0.92\linewidth]{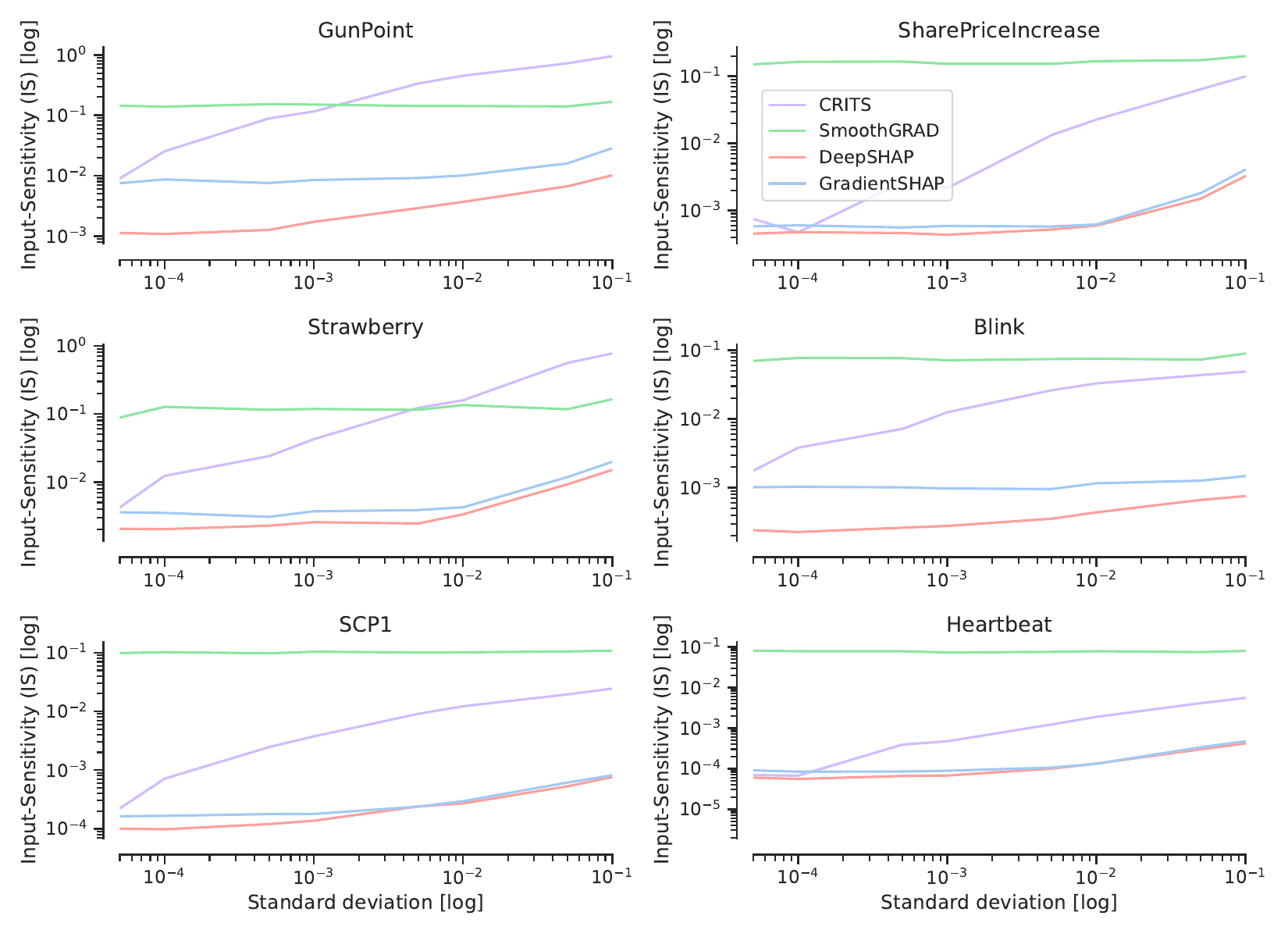}
    \caption{The input-sensitivity of the explanations for each datasets.}
    \label{fig:input-sensitivity}
\end{figure}

Figure \ref{fig:input-sensitivity} shows the plot of the input-sensitivity of the explanations. Ideally, the explanations should not considerably change if the input is slightly changed, and they should do so if the input is significantly changed.

The input-sensitivity of CRITS shows an increasing behavior as the noise increases. With the initial standard deviation noise, the IS of the CRITS explanations is at least $0.0000001$ and with the largest noise it is close to $1$. In contrast, SmoothGRAD presents an IS of $0.1$, and almost no sensitivity to the noise increase. This may be explained in the way SmoothGRAD obtains the explanations: since it originally perturbs the instances and averages the extracted weights throughout these perturbations, adding more perturbations does not further alter the explanations. This means that SmoothGRAD, although providing stable explanations with respect to noise, may not be able to differentiate the feature relevance of instances that could be considerably different, i.e., the local explanations of instances that are dissimilar may be similar and vice versa.

With the largest noise, the standard deviation corresponds to 10$\%$ of the magnitude of the normalized values, which is a considerable amount. The DeepSHAP and GradientSHAP methods also present low explanation sensitivity when the noise magnitude is low, but they do not significantly increase as the noise increases, even when reaching the standard deviation of $10^{-1}$, which may also indicate a lack of differentiation of feature weights among dissimilar instances. 

\subsubsection{Sparsity.} 
\begin{figure}[t!]
    \centering
    \includegraphics[trim={0.15cm 0.2cm 0.4cm 0cm},clip,width=0.91\linewidth]{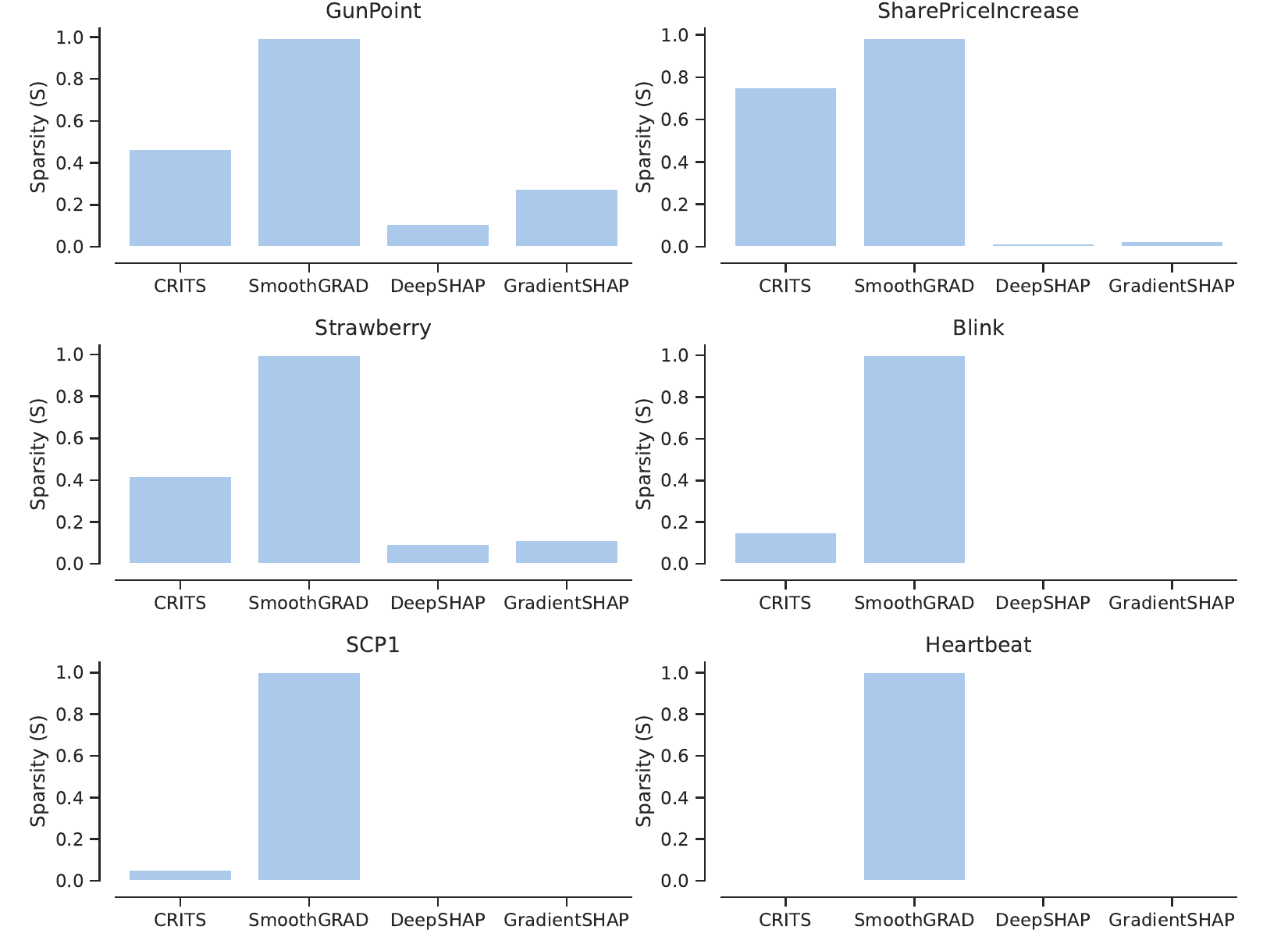}
    \caption{The sparsity of the explanations for each dataset and method.}
    \label{fig:sparsity}
\end{figure}

Figure \ref{fig:sparsity} shows the sparsity of the explanations for each of the explanations methods and datasets. The lower the sparsity, the lower the number of relevant features and the better the explanations in terms of simplicity and understandability. SmoothGRAD is notably the worst performer in sparsity since most of the input features are assigned a relatively high weight through the SmoothGRAD process, which smooths out highly relevant feature weights with its surrounding features. The other methods, including the explanations from CRITS, show a lower number of relevant features.

\section{Conclusions and future work}
\label{sec:conclusions}

CRITS is an intrinsically interpretable TS classifier that can deliver high classification performance and local explanations in UTS and MTS. We have demonstrated its capabilities on different datasets and compared its performance in terms of classification with the ROCKET algorithm, which is one of the best TS classifiers. Additionally, we have extended the unwrapper to work for convolutional-based classifiers to be able to attain intrinsic interpretability. We compared the intrinsic explanations extracted from CRITS, which are the exact weights used by the model to classify each of the instances, to those extracted by state-of-the-art post-hoc explanation methods applied on the CRITS architecture, and showed that the explanations from CRITS have a high alignment, good input-sensitivity, and relatively good sparsity in the datasets tested. We are aware that additional work is needed to generalize CRITS in a broader set of test cases, including potential modifications to CRITS to make it more accurate in complex scenarios, without sacrificing its intrinsic interpretability.

\section{Acknowledgments}
This work was supported by Spanish National Research Agency  (MCIN/AEI/ 10.13039/501100011033) under the Grant PID2020-115401GB-I00.

\bibliographystyle{splncs04}
\bibliography{bibliography}

\begin{thebibliography}{10}
\providecommand{\url}[1]{\texttt{#1}}
\providecommand{\urlprefix}{URL }
\providecommand{\doi}[1]{https://doi.org/#1}

\bibitem{biran2017explanation}
Biran, O., Cotton, C.: Explanation and justification in machine learning: A
  survey. In: IJCAI-17 workshop on explainable AI (XAI). vol.~8, pp. 8--13
  (2017)

\bibitem{burstein2019proceedings}
Burstein, J., Doran, C., Solorio, T.: Proceedings of the 2019 conference of the
  north american chapter of the association for computational linguistics:
  Human language technologies, volume 1 (long and short papers). In:
  Proceedings of the 2019 Conference of the North American Chapter of the
  Association for Computational Linguistics: Human Language Technologies,
  Volume 1 (Long and Short Papers) (2019)

\bibitem{chollet2021deep}
Chollet, F.: Deep learning with Python. Simon and Schuster (2021)

\bibitem{dempster2020rocket}
Dempster, A., Petitjean, F., Webb, G.I.: Rocket: exceptionally fast and
  accurate time series classification using random convolutional kernels. Data
  Mining and Knowledge Discovery  \textbf{34}(5),  1454--1495 (2020)

\bibitem{dorle2020learning}
Dorle, A., Li, F., Song, W., Li, S.: Learning discriminative virtual sequences
  for time series classification. In: Proceedings of the 29th ACM International
  Conference on Information \& Knowledge Management. pp. 2001--2004 (2020)

\bibitem{gupta2020approaches}
Gupta, A., Gupta, H.P., Biswas, B., Dutta, T.: Approaches and applications of
  early classification of time series: A review. IEEE Transactions on
  Artificial Intelligence  \textbf{1}(1),  47--61 (2020)

\bibitem{hsu2019multivariate}
Hsu, E.Y., Liu, C.L., Tseng, V.S.: Multivariate time series early
  classification with interpretability using deep learning and attention
  mechanism. In: Advances in Knowledge Discovery and Data Mining: 23rd
  Pacific-Asia Conference, PAKDD 2019, Macau, China, April 14-17, 2019,
  Proceedings, Part III 23. pp. 541--553. Springer (2019)

\bibitem{ismail2019deep}
Ismail~Fawaz, H., Forestier, G., Weber, J., Idoumghar, L., Muller, P.A.: Deep
  learning for time series classification: a review. Data mining and knowledge
  discovery  \textbf{33}(4),  917--963 (2019)

\bibitem{jia2019improving}
Jia, Y., Bailey, J., Ramamohanarao, K., Leckie, C., Houle, M.E.: Improving the
  quality of explanations with local embedding perturbations. In: Proceedings
  of the 25th ACM SIGKDD International conference on knowledge discovery \&
  Data Mining. pp. 875--884 (2019)

\bibitem{kiangala2020effective}
Kiangala, K.S., Wang, Z.: An effective predictive maintenance framework for
  conveyor motors using dual time-series imaging and convolutional neural
  network in an industry 4.0 environment. Ieee Access  \textbf{8},
  121033--121049 (2020)

\bibitem{kuratomi2020prediction}
Kuratomi, A., Lindgren, T., Papapetrou, P.: Prediction of global navigation
  satellite system positioning errors with guarantees. In: Joint European
  Conference on Machine Learning and Knowledge Discovery in Databases. pp.
  562--578. Springer (2020)

\bibitem{kuratomi2022juice}
Kuratomi, A., Miliou, I., Lee, Z., Lindgren, T., Papapetrou, P.: Juice:
  Justified counterfactual explanations. In: Discovery Science: 25th
  International Conference, DS 2022, Montpellier, France, October 10--12, 2022,
  Proceedings. pp. 493--508. Springer (2022)

\bibitem{lundberg2017unified}
Lundberg, S.M., Lee, S.I.: A unified approach to interpreting model
  predictions. Advances in neural information processing systems  \textbf{30}
  (2017)

\bibitem{maletzke2013time}
Maletzke, A.G., Lee, H.D., Batista, G.E., Rezende, S.O., Machado, R.B.,
  Voltolini, R.F., Maciel, J.N., Silva, F.: Time series classification using
  motifs and characteristics extraction: a case study on ecg databases. In:
  Fourth International Workshop on Knowledge Discovery, Knowledge Management
  and Decision Support. pp. 322--329. Atlantis Press (2013)

\bibitem{miller2019explanation}
Miller, T.: Explanation in artificial intelligence: Insights from the social
  sciences. Artificial intelligence  \textbf{267},  1--38 (2019)

\bibitem{molnar}
Molnar, C.: Interpretable machine learning: A guide for making black-box models
  explainable (2021),
  \url{https://christophm.github.io/interpretable-ml-book/limo.html}

\bibitem{ribeiro2016should}
Ribeiro, M.T., Singh, S., Guestrin, C.: "why should i trust you?" explaining
  the predictions of any classifier. In: Proceedings of the 22nd ACM SIGKDD
  international conference on knowledge discovery and data mining. pp.
  1135--1144 (2016)

\bibitem{rojat2021explainable}
Rojat, T., Puget, R., Filliat, D., Del~Ser, J., Gelin, R.,
  D{\'\i}az-Rodr{\'\i}guez, N.: Explainable artificial intelligence (xai) on
  timeseries data: A survey. arXiv preprint arXiv:2104.00950  (2021)

\bibitem{rudin_2}
Rudin, C.: Stop explaining black box machine learning models for high stakes
  decisions and use interpretable models instead. Nature Machine Intelligence
  \textbf{1}(5),  206--215 (2019)

\bibitem{ruiz2021great}
Ruiz, A.P., Flynn, M., Large, J., Middlehurst, M., Bagnall, A.: The great
  multivariate time series classification bake off: a review and experimental
  evaluation of recent algorithmic advances. Data Mining and Knowledge
  Discovery  \textbf{35}(2),  401--449 (2021)

\bibitem{schlegel2019towards}
Schlegel, U., Arnout, H., El-Assady, M., Oelke, D., Keim, D.A.: Towards a
  rigorous evaluation of xai methods on time series. In: 2019 IEEE/CVF
  International Conference on Computer Vision Workshop (ICCVW). pp. 4197--4201.
  IEEE (2019)

\bibitem{selvaraju2017grad}
Selvaraju, R.R., Cogswell, M., Das, A., Vedantam, R., Parikh, D., Batra, D.:
  Grad-cam: Visual explanations from deep networks via gradient-based
  localization. In: Proceedings of the IEEE international conference on
  computer vision. pp. 618--626 (2017)

\bibitem{shrikumar2017learning}
Shrikumar, A., Greenside, P., Kundaje, A.: Learning important features through
  propagating activation differences. In: International conference on machine
  learning. pp. 3145--3153. PMLR (2017)

\bibitem{smilkov2017smoothgrad}
Smilkov, D., Thorat, N., Kim, B., Vi{\'e}gas, F., Wattenberg, M.: Smoothgrad:
  removing noise by adding noise. arXiv preprint arXiv:1706.03825  (2017)

\bibitem{sudjianto2020unwrapping}
Sudjianto, A., Knauth, W., Singh, R., Yang, Z., Zhang, A.: Unwrapping the black
  box of deep relu networks: interpretability, diagnostics, and simplification.
  arXiv preprint arXiv:2011.04041  (2020)

\bibitem{theissler2022explainable}
Theissler, A., Spinnato, F., Schlegel, U., Guidotti, R.: Explainable ai for
  time series classification: A review, taxonomy and research directions. IEEE
  Access  (2022)

\bibitem{veerappa2022validation}
Veerappa, M., Anneken, M., Burkart, N., Huber, M.F.: Validation of xai
  explanations for multivariate time series classification in the maritime
  domain. Journal of Computational Science  \textbf{58},  101539 (2022)

\bibitem{widdicombe2021gradient}
Widdicombe, A., Julier, S.J.: Gradient-based interpretability methods and
  binarized neural networks. arXiv preprint arXiv:2106.12569  (2021)

\bibitem{zeiler2014visualizing}
Zeiler, M.D., Fergus, R.: Visualizing and understanding convolutional networks.
  In: Computer Vision--ECCV 2014: 13th European Conference, Zurich,
  Switzerland, September 6-12, 2014, Proceedings, Part I 13. pp. 818--833.
  Springer (2014)

\bibitem{zhang2020tapnet}
Zhang, X., Gao, Y., Lin, J., Lu, C.T.: Tapnet: Multivariate time series
  classification with attentional prototypical network. In: Proceedings of the
  AAAI Conference on Artificial Intelligence. vol.~34, pp. 6845--6852 (2020)

\bibitem{zhou2016learning}
Zhou, B., Khosla, A., Lapedriza, A., Oliva, A., Torralba, A.: Learning deep
  features for discriminative localization. In: Proceedings of the IEEE
  conference on computer vision and pattern recognition. pp. 2921--2929 (2016)

\bibitem{zhou2021salience}
Zhou, L., Ma, C., Shi, X., Zhang, D., Li, W., Wu, L.: Salience-cam: Visual
  explanations from convolutional neural networks via salience score. In: 2021
  International Joint Conference on Neural Networks (IJCNN). pp.~1--8. IEEE
  (2021)

\end{thebibliography}

\newpage

\appendix

\section{Hyperparameter tuning} \label{appendix:hyperparameters}

We run a random search for the selection of the CRITS models parameters. This architecture has three main components: (1) the single convolutional layer, (2) a global max-pooling layer and (3) a rectifier network before the output layer.

The convolutional layer component has two main parameters, namely: the size of each kernel and the number of kernels or filters to apply. The max-pooling layer does not have parameters to tune. The rectifier network is a deep neural network with ReLU activations, and therefore has the number of layers and the number of neurons per layer to define as parameters. The activation functions are forced to be ReLU to preserve the local interpretability. All the configurations tried in the search space use the Adam solver.

Each of the parameters has a search range in the random search process, defined as shown in Table \ref{tab:hyperparameter_search_space}.

\begin{table}[]
    \centering
    \begin{tabular}{l|cc}
        \textbf{Component} & \textbf{Parameter} & \textbf{Search space} \\
        \midrule
        \begin{tabular}{l} Convolutional \\ layer \end{tabular} & \begin{tabular}{c} Kernel size \\ \midrule \# kernels \end{tabular} & \begin{tabular}{c} $[3, 16]$ (step 1) \\ \midrule $[16, 256]$ (step: 16) \end{tabular} \\
        \midrule
        \begin{tabular}{l} Rectifier \\ network \end{tabular} & \begin{tabular}{c} \# of layers \\ \midrule Neurons \end{tabular} & \begin{tabular}{c} $[3, 5]$ (step: 1) \\ \midrule $[20, 200]$ (step: 10) \end{tabular} \\
        \bottomrule
    \end{tabular}
    \caption{Hyperparameter search space for the CRITS architecture.}
    \label{tab:hyperparameter_search_space}
\end{table}

The hyperparameter search is done using a and 80$\%$/20$\%$ split for training and testing. A total of 500 samples from the search space, and 2 trials per sample. The final best score in the test set is obtained per dataset, as well as the final configuration of the CRITS architecture classifier. These results are shown in Table \ref{tab:random_search_results}.

\begin{table}[]
    \centering
    \begin{tabular}{l|ccc}
        \textbf{Dataset} & \textbf{Convolutional layer} & \textbf{Rectifier network} & \textbf{F1-score} \\ 
        \midrule
        GunPoint & Kernel: $16 \times m$, \# kernels: 144 & (160, 120, 40, 30, 190) & 0.99 \\
        \midrule
        SharePriceIncrease & Kernel: $16 \times m$, \# kernels: 256 & (180, 120, 180, 130, 140) & 0.62 \\
        \midrule
        Strawberry & Kernel: $16 \times m$, \# kernels: 240 & (120, 200, 180, 190, 20) & 0.99 \\
        \midrule
        Blink & Kernel: $30\times m$, \# kernels: 32 & (200, 50, 200) & 1.00 \\
        \midrule
        SCP1 & Kernel: $15\times m$, \# kernels: 48 & (70, 150, 60, 120, 170) & 0.88 \\
        \midrule
        Heartbeat & Kernel: $8\times m$, \# kernels: 176 & (190, 190, 90, 180, 190) & 0.72 \\
        \bottomrule
    \end{tabular}
    \caption{Best hyperparameters for the CRITS architecture for each dataset.}
    \label{tab:random_search_results}
\end{table}

\section{Explanations saliency maps} \label{appendix:qualitative_results}

The following figures illustrate the relevance obtained by each of the explainers methods under study on the CRITS classifier architecture. 

\begin{figure}
    \centering
    \includegraphics[trim={0.2cm 0.25cm 0.25cm 0.25cm},clip,width=0.95\linewidth]{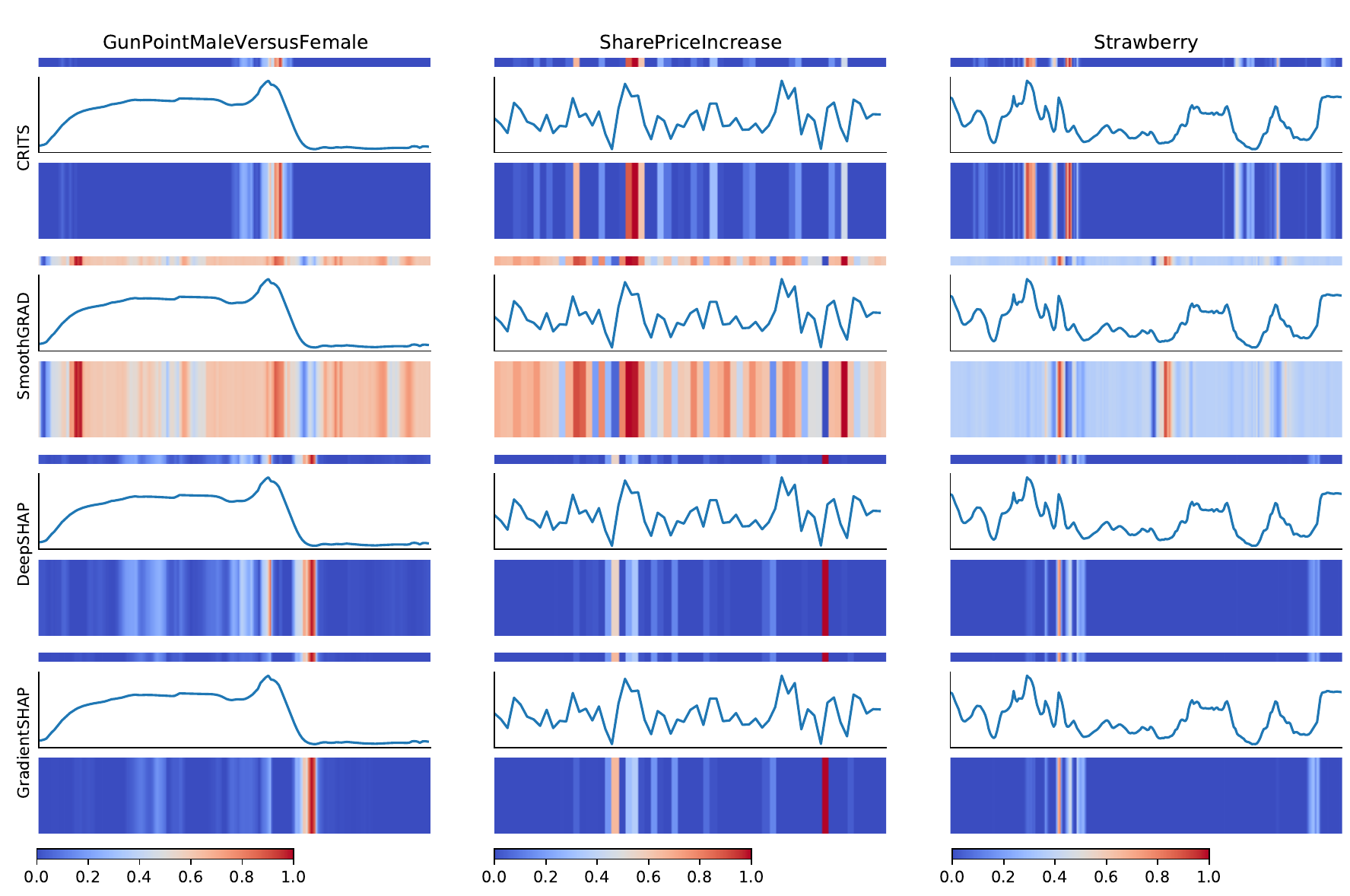}
    \caption{Relevance values, $\mathbf{r}^i$, for the univariate datasets.}
    \label{fig:crits_relevances}
\end{figure}

\begin{figure}
    \centering
    \includegraphics[trim={0.2cm 0.25cm 0.25cm 0.25cm},clip,width=0.95\linewidth]{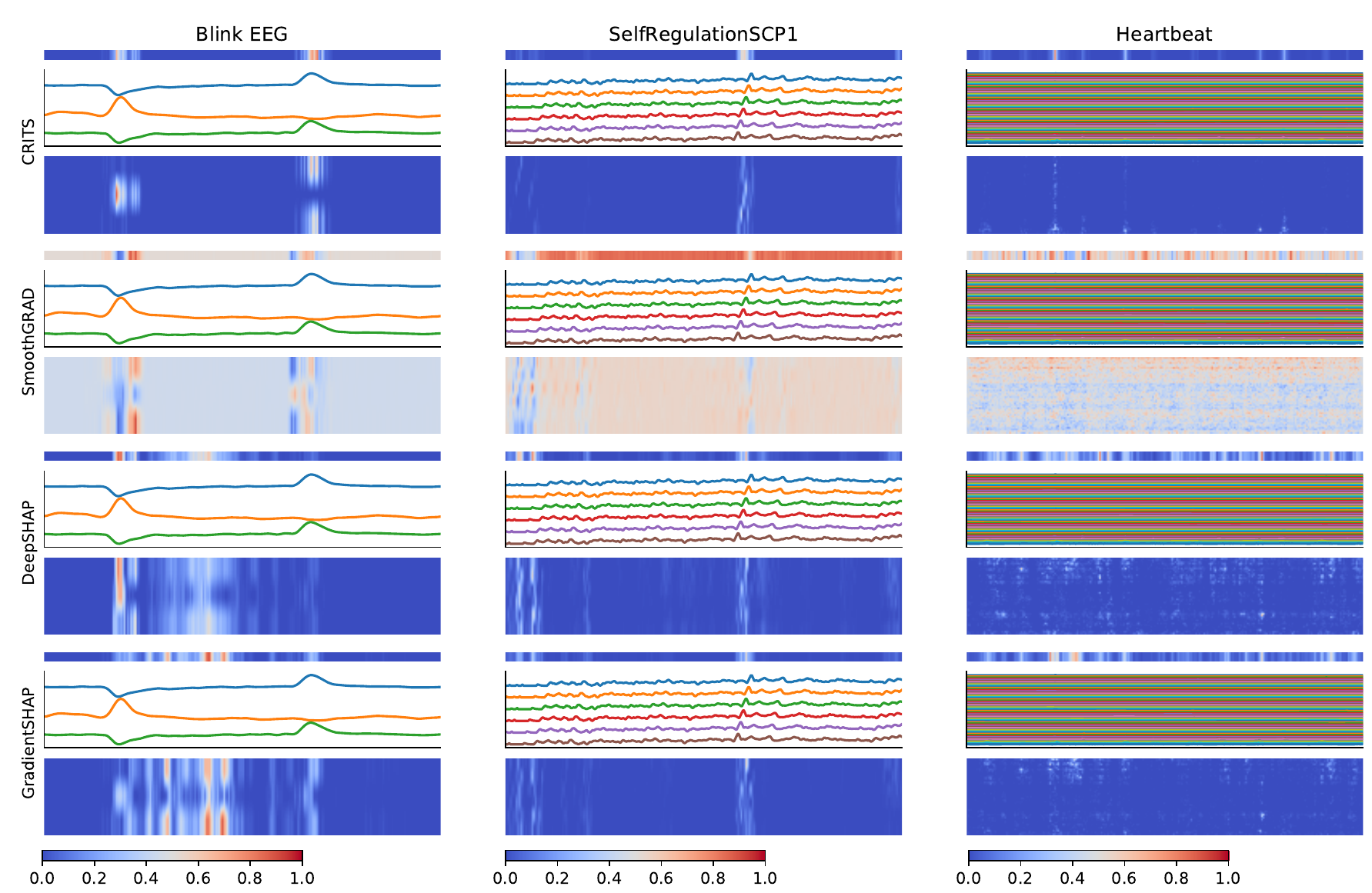}
    \caption{Relevance values, $\mathbf{r}^i$, for the multivariate datasets.}
    \label{fig:crits_relevances}
\end{figure}
\end{document}